# Comparison of the C4.5 and a Naive Bayes Classifier for the Prediction of Lung Cancer Survivability

George Dimitoglou, James A. Adams, and Carol M. Jim

**Abstract**—Numerous data mining techniques have been developed to extract information and identify patterns and predict trends from large data sets. In this study, two classification techniques, the J48 implementation of the C4.5 algorithm and a Naive Bayes classifier are applied to predict lung cancer survivability from an extensive data set with fifteen years of patient records. The purpose of the project is to verify the predictive effectiveness of the two techniques on real, historical data. Besides the performance outcome that renders J48 marginally better than the Naive Bayes technique, there is a detailed description of the data and the required pre-processing activities. The performance results confirm expectations while some of the issues that appeared during experimentation, underscore the value of having domain-specific understanding to leverage any domain-specific characteristics inherent in the data.

──────────────────────

- *George Dimitoglou is with the Department of Computer Science at Hood College, Frederick, MD 21701, USA. E-mail: dimitoglou@cs.hood.edu*
- *James A. Adams is with Marriott International, Bethesda, MD 20817, USA.*
- *Carol M. Jim is with the School of Engineering & Applied Science at The George Washington University, Washington, DC 20052, USA.*

────────────── ◆ ──────────────

## 1 INTRODUCTION

The proliferation of computing technologies in every aspect of modern life, has substantially increased the volume of data being collected and stored. The availability of these massive data sets introduces the challenge of having to analyze, understand and use the data. Using data mining techniques, data can be classified to bring forward interesting patterns or serve as a predictor for future trends. Numerous algorithmic techniques have been developed to extract information and discern patterns that may be useful for decision support. Such techniques are routinely applied in various areas and problems ranging from business, marketing and money fraud detection, to security surveillance, scientific research and health care.

This paper uses the Surveillance, Epidemiology and End Results (SEER) dataset [24], to compare the predictive effectiveness of two algorithmic techniques. It uses lung cancer survivability as the objective criterion to assess how well each technique is able to predict patient survivability given a set of historical data. This type of analysis is not unique, in fact it was inspired by a similar analysis on breast cancer survivability techniques [2].

The particular disease was selected because of its severity. Lung cancer is one of the most prevalent and the most deadly type of cancer. It manifests as an uncontrolled cell growth in lung tissues. Both sexes are susceptible, with men having higher probability (1 in 13) of developing the disease in a lifetime than women (1 in 16). These probabilities include smokers and non-smokers alike with smokers being at a much higher risk. Smoking has been identified as the greatest contributing cause [3, 12, 22] but exposure to asbestos [6, 16] and radon gas [5] have also been linked with the incidence of the disease.

In the United States, while lung cancer diagnosis comes second to prostate cancer for men and breast cancer for women, it is the leading cause of cancer death across the sexes. In 2006, lung cancer diagnoses accounted for 196,454 patients of all new cancer patients and lung cancer mortality accounted for approximately 29% of all cancer-related deaths (158,599 patients) [25]. These mortality rates remained stable in 2010 indicating that more people die of lung cancer than of colon, breast, and prostate cancer combined [1].

Developing treatments and better understanding the characteristics of the disease is almost exclusively based on clinical and biological research. However, the abundance of cancer research related data provides new opportunities to develop data and statistics-driven that may complement the traditional research approaches.

One of the most interesting and challenging areas is *survivability* or *disease outcome prediction*. This area lends itself well for analysis using data mining techniques. With survival analysis, disease-specific historical data is examined and analyzed over a certain period of time to develop a prognosis.

The increased interest in health informatics to provide more effective health care and the abundance of computing power, massive data storage capacity and the development of automated tools, has allowed the collection of large volumes of medical data. As a result, methods and techniques from the computing sciences related to knowledge discovery and data mining have become very useful in identifying patterns and analyzing relationships in the data that may lead to better disease outcome prediction [7, 8, 23, 26]. This type of pattern analysis can be used in the development of patient survivability prediction models. It is the prevalence and serious effects of lung cancer and how well existing data lends itself for data mining that provided the motivation to select this subject as the platform for the comparison of the two algorithmic techniques.

One of the special aspects of our study is the volume and authenticity of the experimentation dataset. We used data provided the SEER program, one of the most comprehensive sources on cancer incidence and survival in the U.S. provided by the National Cancer Institute. For the analysis, we used two different classification algorithms: the Naive

Bayes and J48, an implementation of the C4.5 algorithm.

The following section contains a brief overview of the relevant algorithms and prediction models. Section 3 provides a detailed description of the data and pre-processing activities. Section 4 describes survivability in the context of the specific SEER dataset and it is followed by Section 5 which presents the experimentation and results. The final sections provide a discussion of the results along with some conclusions and areas for future work.

## 2 ALGORITHMS AND PREDICTION MODELS

To develop a prediction model, classifiers are the data mining algorithm of choice. The basic premise is based on treating a collection of cases as input, each belonging to a small number of classes described by a fixed set of attributes [17]. The classifier's output is the prediction of the class to which a new case belongs. The accuracy of the prediction varies depending on the classifier and the types of attributes and classes in the dataset. Classification may be viewed as mapping from a set of attributes to a particular class [17]. For the data analysis, two classification techniques were used, the Naive Bayes and the open source version of the C4.5 statistical classifier.

### 2.1 Naive Bayes

The Naive Bayes algorithm is a simple probabilistic classifier that calculates a set of probabilities by counting the frequency and combinations of values in a given data set.

The probability of a specific feature in the data appears as a member in the set of probabilities and is derived by calculating the frequency of each feature value within a class of a training data set. The training dataset is a subset, used to train a classifier algorithm by using known values to predict future, unknown values.

The algorithm uses Bayes theorem [15] and assumes all attributes to be independent given the value of the class variable. This conditional independence assumption rarely holds true in real-world applications, hence the characterization as Naive yet the algorithm tends to perform well and learn rapidly in various supervised classification problems [18].

This "naivety" allows the algorithm to easily construct classifications out of large data sets without resorting to complicated iterative parameter estimation schemes.

### 2.2 J48

J48 is an implementation of C4.5 [20] which is the successor of the ID3 algorithm [19]. ID3 is based on inductive logic programming methods, constructing a decision tree based on a training set of data and using an entropy measure to determine which features of the training cases are important to populate the leaves of the tree.

The algorithm first identifies the dominant attribute of the training set and sets it as the root of the tree. Second, it creates a leaf for each of the possible values the root can take. Then, for each of the leaves it repeats the process using the training set data classified by this leaf [18]. The core function of the algorithm is determining the most appropriate attribute to best partition the data into various classes. The ID3 algorithm [21] uses entropy and information gain [29], borrowed from information theory, to measure the impurity of the data items.

Smaller entropy values indicate full membership of the data to a single class while higher entropy values indicate existence of more classes the data should belong to. Information gain [29] measures the decrease of the weighted average entropy of the attributes compared with the entropy of the complete dataset. Therefore, the attributes with higher information gain are better classification candidates for data items. One limitation of ID3 is its sensitivity to features with a large population of values. C4.5, an ID3 extension, was developed to address this limitation by finding high accuracy hypotheses, based on pruning the rules issued from the constructed tree during the leaf population phase. However, C4.5 is computationally more intensive, in terms of time and space complexity.

The comparison of the two techniques in the literature, Naive Bayes and J48, does not render conclusive results. Various studies in different domains have found each technique performing differently than the other. For example, in the context of intrusion detection systems, one study [10] found J48 performing better than another similar study [16], while studies in the prediction of acute cystitis and nephritis [14] and the classification of arrhythmia data [23] the performance of Naive Bayes was superior. Similarly, in the context of spam e-mail identification [27], J48 performed better in terms of accuracy of overall classification despite the widespread use of Bayesian-based span filtering techniques.

## 3 DATA

### 3.1 Data Description

The data used in this study was acquired from the SEER database. We used version 3.4.10 of the open source toolkit Weka to Environment for Knowledge Analysis (WEKA) [13] to determine the accuracy of two algorithms in predicting lung cancer survivability. The data used in the experiments is sourced from the RESPIR incidence data files included in three SEER datasets (Table 1).

TABLE 1
SEER DATASETS AND YEARS OF COVERAGE.

| Dataset Name | Range | Years |
|---|---|---|
| yr1973 2004.seer9 | 1973-2004 | 31 |
| yr1992 2004.sj la rg ak | 1992-2004 | 12 |
| yr2000 2004.ca ky lo nj | 2000-2004 | 4 |

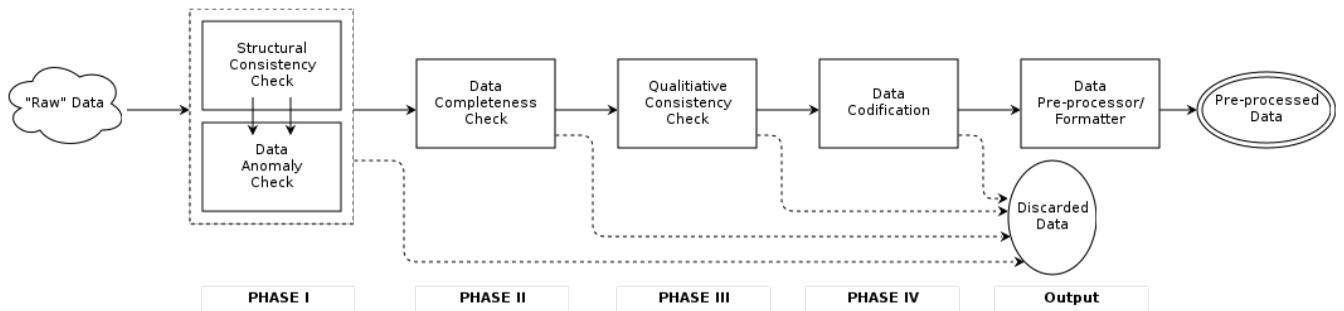

Fig. 1. Diagram of the four phases of pre-processing activities. Starting from the "raw data" a series of data integrity and consistency checks prepare the data for mining.

There is a total of 481,432 records in the dataset. It contains records of patients diagnosed between 1973-2004. However, due to the modification and addition of data fields, records prior to 1988 were removed for our study.

The remaining records were filtered to include only those that contain the Extent of Disease (EOD) 10-digit coding system. This coding system provides detailed, relevant and useful information, such as tumor size, number of positive nodes, number of primaries and other quantifiers that greatly enrich the dataset. The effect of this filter was to restrict the dataset to records containing diagnoses only between the complete years in the dataset (1988-2003).

### 3.2 Pre-processing Activities

The records in the dataset are not disease-specific so they include multiple cancer types and disease profiles. To prepare the data for mining numerous pre-processing activities were performed. Figure 1 depicts the pre-processing workflow. It is divided in four interconnected stages with the overall objective to ensure that data is appropriate, complete and well formatted.
The input to the pre-processing workflow is the source, unprocessed, "raw", data files.

The first phase (Phase I) comprises of two activities. The first activity is to perform a structural consistency check to ensure that field names and record structures are homogeneous between different source files. It is not uncommon for different year data sets to have different fields due to changes in the data collection methodology. The second activity is to identify anomalous data elements such as missing values and erroneous data. For example, empty fields may often be denoted by leaving the field blank, labeling it as "*null*" or having a designation such as "*N/A*" or "*-0-*". While any label to indicate a missing value is acceptable, the convention followed in the data must be identified and confirmed.

Phase II examines the data with respect to relevancy and completeness. Checking for relevancy ensures that pertinent information is included to accomplish the study objective. For example, data not containing any explicit information about the patient's type of cancer.

In this case, the data has to be further processed to determine if a synthesis of multiple fields can help to derive and substitute the missing information. If it is not possible, the dataset may be deemed inappropriate and be discarded. Another result from this phase is to ensure that data granularity is sufficient for the study objective. If the objective is to identify individual patient-based relationships but the collection only provides information at the group level, then the data would not have the necessary granularity and should be discarded.

In the third phase, the data is examined for qualitative consistency. The objective is to ensure the dataset as a whole is meaningful and useful. For example, when mining for the onset time of prostate cancer, a type of cancer appearing exclusively on male patients, qualitative consistency may indicate that the data includes female patients. Such inclusion pollutes the dataset and will most likely skew the results. Similarly, extraneous ages, such as patients over 120 years of age or younger than 0 years of age, would be indicators of questionable entries in the dataset.

The fourth phase requires data fields to be codified using different types of values (ex. ordinal, categorical, etc.). Some of the fields can be immediately used for analysis while others must be normalized to provide more meaningful numeric ranges or to better identify described conditions. The final pre-processing task is related to expected format and syntax modifications required by the data mining algorithms and software.

In each of the four stages, it is possible that some data may be deemed incomplete or too coarse for mining. In such cases the data may be discarded, unless the discarded set significantly impacts the integrity of the dataset. It is also possible that the mining algorithms can handle anomalous conditions by labeling or ignoring certain data.

### 3.3 Metadata

The SEER data is well-described providing item descriptions, codes, and coding instructions. However, the dataset includes incidence records for all possible cancer types, treatments and death causes along with a wealth of non-identifiable patient information.

After the data is pre-processed, certain data elements present themselves with interesting information while others become candidates for omission. In the next two sections, a brief description of the data elements is provided, followed by a post-processing filtering.

While this process is iterative after each pre-processing phase, here we present the final result of the end-to-end pre-processing.

The following data elements were included in the SEER dataset:

**Age**. Patient age at initial lung cancer diagnosis.

**Race**. Patient race described using one of seven categories: *White*, *Black*, *American Indian/Alaskan Native*, *Asian* or *Pacific Islander*, *Other Unspecified* and *Unknown*.

**Marital Status**. Patient marital status (*Single*, *Married*, *Separated*, *Divorced*, *Widowed* and *Unknown*) at the time of diagnosis for the reportable tumor.

**Primary Site Code**. Tumor location in the body, as classified by the International Classification of Diseases for Oncology (ICD-O-3) [9] for topography codes.

**Histologic Type**. Tumor morphology using 178 distinct ICD-O-3 codes.

**Behavior Code**. Indicates if tumor is malignant.

**Tumor Size**. Indicates the tumor size.

**Grade**. Subjective description of tumor's appearance between visits.

**Extension of Tumor**. Farthest documented extension of tumor away from the primary site.

**Lymph Node Involvement**. (Binary value). Indicates if the tumor involves Lymph Node chains.

**Site Specific Surgery Code**. Describes body and organ locations that could have been operated on.

**Radiation**. Radiation therapy method on first treatment.

**Stage of Cancer**. Physical location and spread.

**Radiation sequence with Surgery**. Sequence of administering radiation such as *pre-surgery*, *post-surgery* and *pre-/post-surgery radiation*.

**Survival Code**. Denotes if patient has survived. A set of "survival codes" was created to identify classes of surviving patients. *Survival Time Recode* expresses the survival time in months, *Vital Status Recode* indicates if patient is alive and *Death Code* indicates the recorded cause of death based on ICD-10 [28].

### 3.4 Filtering

Mining algorithms assume the data to be noise-free. The output of particularly the third pre-processing phase (qualitative consistency checks), reveals a number of interesting characteristics in the data.

For example, there are two fields that are eliminated from the dataset. The *Behavior Code* indicates if a tumor is malignant. In our dataset, the majority (99.9%) of the records are identified as malignant and the remaining (00.1%) contain a *null* value. This field is not meaningful for our particular study because unlike other cancer types, lung cancers, are assumed to be malignant for all cases. Similarly, it is questionable how meaningful the *Grade* field is. It holds a codified, subjective description of the tumor's appearance between physician visits. While there is ordering of the possible values (Table 2) yet it is unknown how much worse (or better) the different grades vary between each other.

A preliminary analysis indicates that almost half (48.36%) of all the data fall in the Grade IV category. Within this category, approximately half of the entries are recorded as not available (*N/A*).

Therefore, the Grade field is discarded because it is populated by subjective observational data. In addition, the unclear demarcation of categories and information unavailability would increase the noise in the data rather provide a meaningful discriminatory attribute.

TABLE 2
THE *GRADE* FIELD WITH THE POSSIBLE RANGE OF VALUES.

| Grade | Possible Tumor Size Descriptions |
|---|---|
| Grade I | Well differentiated<br>Differentiated |
| Grade II | Moderately differentiated<br>Well differentiated |
| Grade III | Poorly differentiated<br>Differentiated |
| Grade IV | Undifferentiated<br>Anaplastic, cell type not determined<br>Not stated<br>Not Available (N/A) |

Two other fields require special attention. The *Primary Site Code* contains 28 distinct codes to describe the incidence of cancer. Not all of the codes are applicable so only the records with codes identifying lung cancer (*C340-C343*, *C348-C349*) are preserved.

The *Age* field on the other hand, has values ranging from 1 to 106. This is a very wide value spread which may generate many, non-significant data points. As the significance of age ordinality becomes a concern, it is probably not very meaningful to differentiate between a 24-year old and a 25-year old patient. To address this, the data is discretized into an 8-binary scale format with the following entries: (0-24), (25-34), (35-44), (45-54), (55-64), (65-74), (75-84), >= 85.

### 3.5 Preliminary Statistical Analysis Results

At this stage a number of statistical measures and counts are applied to identify if there are any interesting patterns simply by measuring the frequency of occurrence of certain elements and events in the data. Early results for the *Tumor Size* keyword indicate a trend of surviving patient records having this attribute with a value of less than 40mm.

Another interesting observation can be made about the *Site Specific Surgery Code* which describes the body and organ locations that could have been operated on. The majority (95%) of the records are identified as *Lung and Bronchus* and the remaining as *Pleura and Trachea*, *Mediastinum*, and *Other* respiratory organs.

On the treatment side, the keyword *Radiation*, indicates the method of radiation therapy performed as part of the first treatment. While there are almost ten possible categories, 92% of the records in the dataset are in categories under D*iagnosed at Autopsy* and *Beam Radiation*. The remaining categories are: *Radioactive Implants*, *Radioisotopes*, *Combination Treatment*, *No Method Specified*, P*atient Refused Therapy*, *Radiation Recommended* and *Unknown If Administered*.

While none of the above measurements can lead to any conclusions about the correlation of tumor size and sur-vivability or, surgery location and treatment options, they offer an alternative perspective on the data. Perhaps these findings could prove useful during an association rule learning experiment either as a validation or re-enforcement mechanism.

## 4 SURVIVABILITY

Survival rates and statistics indicate the percentage of patients who manage to survive from cancer for a specific amount of time, typically five years (60 months). Survival must be determined as cause-specific survival, representing survival from lung cancer and not any other cause.

The ICD-10 *Death Code* designation plays a significant role in this determination as patient records indicating failure to survive due to causes other than lung cancer must be eliminated. This assumes the data contains a single, accurate, disease-specific cause of death code. It is often hard to depend on such variable particularly with cancer patients, as the cause of death may often be attributed to a different site due to to metastasis, instead of the primary site. In addition, the survival rates are often vague in specifying whether a patient is still undergoing treatment and if the patient has achieved full or partial remission.

ALGOTITHM 1
DETERMINING SURVIVAL STATUS.

```
1   DeathCode ⇐ (from ICD – 10 designation)
2   if (SurvivalTimeRecode >= 60 months) ∧
3       (VitalStatusRecode = "alive" ) then
4           Survived ⇐ true;
5   else if (SurvivalTimeRecode < 60 months) ∧
6           (cause ∈ DeathCode) then
7           Survived ⇐ false;
8   else if SurvivalCode = null then
9           remove record;
10  else {Survived = inconclusive;}
11          remove record;
12  end if
```

It is therefore necessary to derive a "survived" flag for each patient record. This flag is derived using Algorithm 1 and taking into account the *Death Code* field with the patient's reported status (*VitalStatusRecode*) and the reported survival time (*SurvivalTimeRecode*).

TABLE 3
PROFILE OF THE "SURVIVED" PATIENT CLASS.

| Period | Status | Number of Patients | % |
|---|---|---|---|
| 1988-1999 | Survived | 14,368 | 8.23% |
| | Not Survived | 160,123 | 91.77% |
| Total | | 174,491 | 100.00% |

Any records that contain a *null* value in the survival code after applying the above algorithm are removed. Table 3 shows the number of instances and the respective percentage of patients who survived after being diagnosed with lung cancer.

### 4.1 Survival Time

There is an interesting pattern in the data when comparing the median survival time (MST) in months from year to year. The MST is very stable from year to year, gradually increasing, as expected. However. between the years 1999 and 2000, there is a dramatic decrease.

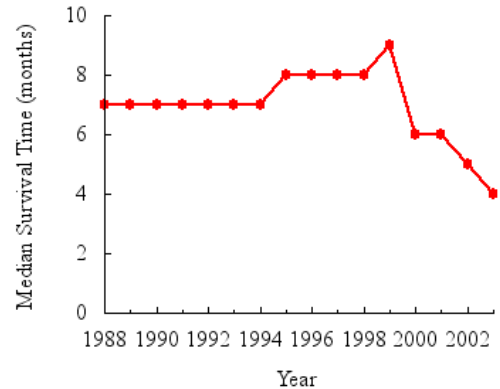

Fig. 2. Median survival time (in months) per year.

The median number of survived months in the 16-year range is constant for the first seven years, gradually increases between year eight and twelve, and then dramatically decreases each year by approximately 33% through the end of the dataset (Figure 2).

The increase in survival rate, which goes against the data in Fig. 2, could be justified with the introduction of a revolutionary treatment and very early prediction after year 2000. Most likely, this decrease is due to a major increase in the number of patient records (Table 4).

After further investigation, we discovered that this decline coincides with coding differences between the multiple SEER reporting sites. In 2000, a new SEER reporting site was introduced, collecting data from densely and highly populated areas of the U.S. including California and New Jersey.

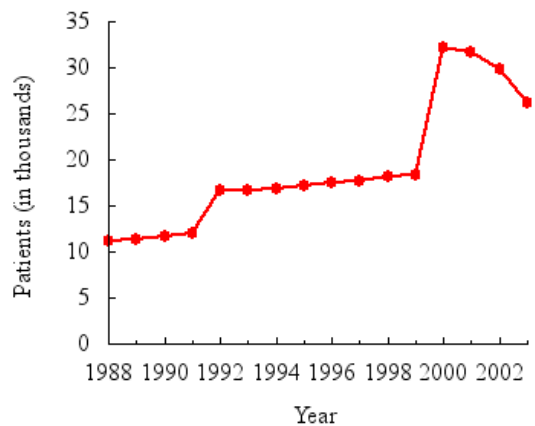

Fig. 3. Number of diagnosed patients (in thousands).

The data from these reporting sites increased the number of records (Table 4) and along with the use of different coding standards may explain the MST decline in the data.

Regional trends or anomalies may also be another contributing factor so further examination is warranted in order to use the specific data segment with confidence. As a result, data after 1999 were discarded from the study.

TABLE 4
MEDIAN SURVIVAL TIME AND NUMBER OF PATIENTS PER YEAR OF DIAGNOSIS (1988-2003).

| Year | Median Survival (months) | Number of Patients |
|---|---|---|
| 1988 | 7 | 11,148 |
| 1989 | 7 | 11,405 |
| 1990 | 7 | 11,726 |
| 1991 | 7 | 12,069 |
| 1992 | 7 | 16,734 |
| 1993 | 7 | 16,764 |
| 1994 | 7 | 16,856 |
| 1995 | 8 | 17,276 |
| 1996 | 8 | 17,530 |
| 1997 | 8 | 17,743 |
| 1998 | 8 | 18,226 |
| 1999 | 9 | 18,395 |
| 2000 | 6 | 32,198 |
| 2001 | 6 | 31,666 |
| 2002 | 5 | 29,959 |
| 2003 | 4 | 26,244 |

Another aspect of survivability is the survival length By setting a limit, patient records can be classified in two survival categories: short-term and long-term. This limit, found by the median survival rate, indicates the number of patients in a year diagnosed within twelve months. To implement this threshold and identify long-term survivability, Rule-1 (Table 5) is used to identify a long-term survivor and classify the patient as a "*survivor*".

Just by applying this rule, 135,878 (30.23%) entries of the 449,446 total patient records were discarded due to insufficient information that forbid us to properly classify patients as *survived* or as *not survived*.

While Rule-1 establishes meaningful constraints for long-term survivability, it is very strict and discriminating for the short and mid-term.

This is because the median survival time is consistently less than a year, and in the early years of the data, it is fairly consistent so that 35-40% of patients survive longer than a year, albeit this number significantly decreases after 2000.

To address these issues, we established additional rules to improve the determination of survivability (Table 5).

TABLE 5
RULES FOR SURVIVABILITY DETERMINATION.

| Rule Number | Expression |
|---|---|
| Rule-1 | *if* patient lives > 60 months *and* (status=alive *or* died after follow-up cutoff date) *then* patient = survivor |
| Rule-2 | *if* no survival time data exists *or* patient diagnosed in last year of our dataset *and* not dead *and* survival < $T_s$ *then* patient record is removed |
| Rule-3 | *if* patient dies *and* cause of death is cancer of lung *or* bronchus *and* survival time < $T_s$ *then* patient= short-term survivor |
| Rule-4 | *if* patient survival time > $T_s$ *then* patient= long-term survivor |
| Rule-5 | *if* none of Rules 2-4 applicable *then* patient record is removed |

## 5  EXPERIMENTATION

The purpose of this study is to test the accuracy of two learning algorithms (Naive Bayes and J48) in predicting survivability of patients diagnosed with lung cancer. After pre-processing, the data is ready to be processed by the two algorithms.

TABLE 6: TRAINING DATA

| Years | Number of Years | Target Year |
|---|---|---|
| 1988-1992 | 5 | 1993 |
| 1992-1998 | 7 | 1999 |
| 1988-1997 | 10 | 1998 |

Three experimentation executions were performed, each under a common scenario: using a variable number of sequential years as a training dataset, attempt to predict the accuracy of lung cancer incidence for a subsequent year (target year).

Then, to determine the level of accuracy for each algorithm, compare the predicted outcomes with the target data. Several datasets are used, identified by the number of years of training and target data (Table 6).

### 5.1 Execution

The Naive Bayes algorithm was executed over each one, five, seven and ten year Training Data date ranges producing correctly classified instances ranging from 88.27% to 92.38% (Table 7). For each experiment, we obtained the Correctly Classified Instances (CCI), the Incorrectly Classified Instances (ICI) the Kappa statistic (κ) and the root-mean-square-error (RMSE). The CCI is the sample accuracy without any chance corrections and it is expressed as a per-centage of correctly classified instances. The Kappa statistic is a chance-corrected measure and reflects the agreement between the classifications and the true classes. Values over zero indicate that the classifier is performing better than chance. The RMSE measures the differences between predicted or estimated values against observed values.

TABLE 7: DETAILED EXPERIMENT RESULTS FOR THE NAIVE BAYES ALGORITHM.

|      | 5 years          | 7 years          | 10 years         |
|------|------------------|------------------|------------------|
| CCI  | 14,577 (92.38%)  | 15,175 (88.27%)  | 15,208 (89.13%)  |
| ICI  | 1,203 (7.62%)    | 2,017 (11.73%)   | 1,855 (10.87%)   |
| κ    | 0.49100          | 0.53610          | 0.51360          |
| RMSE | 0.24710          | 0.30750          | 0.29770          |
| Total| 15,780           | 17,192           | 17,063           |

The J48 algorithm was also executed over the same Training Data date ranges and produced correctly classified instances ranging from 89.63% to 94.44% (Table 8).

TABLE 8: DETAILED EXPERIMENT RESULTS FOR THE J48 ALGORITHM.

|      | 5 years          | 7 years          | 10 years         |
|------|------------------|------------------|------------------|
| CCI  | 14,902 (94.44%)  | 15,409 (89.63%)  | 15,461 (90.61%)  |
| ICI  | 878 (5.56%)      | 2,017 (11.73%)   | 1,855 (10.87%)   |
| κ    | 0.38360          | 0.45090          | 0.40210          |
| RMSE | 0.20970          | 0.29030          | 0.27280          |
| Total| 15,780           | 17,192           | 17,063           |

Comparing the overall accuracy of the two algorithms, it seems that both algorithms performed fairly well, achieving approximately a 90% correct classification rate. The J48 algorithm consistently performed slightly better than the Naive Bayes algorithm in successfully predicting the correct survivability outcome (Table 9).

TABLE 9: ACCURACY COMPARISON BETWEEN THE TWO ALGORITHMS PER EXPERIMENT.

| ID  | Prediction for year | Years of training | Naive Bayes | J48      |
|-----|---------------------|-------------------|-------------|----------|
| T93 | 1993                | 5                 | 92.3764%    | 94.4360% |
| T99 | 1999                | 7                 | 88.2678%    | 89.6289% |
| T98 | 1998                | 10                | 89.1285%    | 90.6113% |

For all executions, the values of the kappa statistic indicate the strength of the classifier's stability and in our experimentation, both algorithms perform well. Similarly, using the RMSE, the low error rate of both algorithms can serve as an indirect measure of effectiveness.

## 6 DISCUSSION

### 6.1 Classification Results

The objective of the experiments was to compare the strength of the two techniques in survivability prediction of patients diagnosed with lung cancer. While an approximately 90% overall correct classification accuracy is achieved, there is an unexpected result.

For both the Naive Bayes and the J48 algorithm, the 5-year training dataset produces a higher accuracy rate than the 7-year and 10-year training datasets. In fact, the 5-year training dataset outperforms and is slightly more accurate than the accuracy provided from the 7- year and 10-year training data sets. It would be expected to have a higher prediction accuracy with more training data but this is not the case. There are two possible factors at play. First, the difference in the number of patient records between the three data sets is not proportional. The 5-year training set includes 15,780 lung cancer patient records and the 7-year training set includes 17,192. This is a difference of 1,412 (an 8.94% increase) of patient records. On the other hand, the 10-year training set includes 17,063 lung cancer patient records, which is a 0.75% decrease (129 patients) than the the 7-year data set. So the difference between the 5-year and 7-year data set is not significantly different. Second, it is likely that the larger training data set may contain inconsistencies due to changes in the data collection processes and locations. The 5-year training set contains data primarily from one SEER reporting location. In 1992, a second SEER reporting location began collecting data, which was included in the bulk of the dataset for the 7- year and 10-year studies.

## 7 CONCLUSION & FUTURE WORK

This study examines the ability of data mining and machine learning methods to accurately predict the survivability of patients diagnosed with lung cancer. Survivability is defined as someone who lives beyond a 5-year period.

Generally, we achieve an approximate prediction accuracy around 90% using either the Naive Bayes or the J48 algorithm. As a result of this study, a treating physician can theoretically collect a handful of medical measurements such as tumor size and location, treatment options, patient's background, etc., and predict with a fairly high degree of accuracy whether the patient is likely to live for five or more years.

Considering the high mortality rate (>90%) of the patients in the study, it seems reasonable and useful to examine the survivability of patients over a shorter time period, for example, between twelve and eighteen months. In light of the findings and observations of the study, future research will be focused on examining regional trends and the impact of location on survivability. More in depth analysis can also be performed by comparing survivability results with data related to regional medical insurance carriers, state laws that affect the medical insurance industry, and other external factors.

In this paper, we compared the effectiveness of the Naive Bayes classification algorithm and the J48 implementation of the C4.5 decision tree algorithm in the prediction of survivability for a specific disease.

From the results, the Naive Bayes technique remained true to its reputation and its portrayal in the literature: it is robust, easy to interpret, it often does surprisingly well but may not be the best classifier in any particular application [15]. In this case, Naive Bayes performed consistently worse than J48. Yet, its simplicity and fairly competitive performance make it an appealing alternative. Overall, both are viable and useful algorithms and the performance differences between them were marginal. More work will be done to investigate if these marginal differences exist due to the way the algorithms are implemented or due to inherent characteristics in the data.

## ACKNOWLEDGMENT

The authors wish to thank SEER and the National Cancer Institute for the data.

## REFERENCES

[1] American Cancer Society, Cancer Facts & Figures. American Cancer Society, Atlanta, GA, 2010, URL: http://www.cancer.org/research/.

[2] Bellaachia, A., Guven, E. Predicting Breast Cancer Survivability using Data Mining Techniques, Ninth Workshop on Mining Scientific and Engineering Datasets in conjunction with the Sixth SIAM International Conference on Data Mining (SDM 2006), Saturday, April 22, 2006.

[3] Biesalski, H K and Bueno de Mesquita, B and Chesson, A and Chytil, F and Grimble, R and Hermus, R J and Kohrle, J and Lotan, R and Norpoth, K and Pastorino, U and Thurnham, D, European Consesus Statement on Lung Cancer: risk factors and prevention. Lung Cancer Panel.. CA Cancer J Clin, 48, 3, 167-76; discussion 164-6, URL: http://www.biomedsearch.com/nih/European-Consensus-Statement-Lung-Cancer/9594919.html.

[4] Catelinois, Olivier and Rogel, Agnes and Laurier, Dominique and Billon, Solenne and Hemon, Denis and Verger, Pierre and Tirmarche, Margot, 2006, Lung cancer attributable to indoor radon exposure in France: impact of the risk models and uncertainty analysis., Environmental Health Perspectives, 114 (9), 1361-6.

[5] Darnton, Andrew J and McElvenny, Damien M and Hodgson, John T, 2006, Estimating the number of asbestos-related lung cancer deaths in Great Britain from 1980 to 2000. The Annals of occupational hygiene, 50 (1), 29-38.

[6] Dechang Chen, Kai Xing, Donald Henson, Li Sheng, Arnold M. Schwartz, and Xiuzhen Cheng. 2008. A Clustering Approach in Developing Prognostic Systems of Cancer Patients. In Proceedings of the 2008 Seventh International Conference on Machine Learning and Applications (ICMLA '08). IEEE Computer Society, Washington, DC, USA, 723-728.

[7] Dursun Delen and Nainish Patil. 2006. Knowledge Extraction from Prostate Cancer Data. In Proceedings of the 39th Annual Hawaii International Conference on System Sciences (HICSS '06), Vol. 5. IEEE Computer Society, Washington, DC, USA, 92.2.

[8] Fritz, A, Jack, A Percy,C Sobin, L Parkin, D.M..(eds.) International classification of diseases for oncology (ICD O). 3d ed. Geneva: World Health Organization, 2000.

[9] Gandhi, G M and Srivatsa, S.K. (2010). Classification Algorithms in Comparing Classifier Categories to Predict the Accuracy of the Network Intrusion Detection - A Machine Learning Approach. Advances in Computational Sciences and Technology, 3 (3), 321- 334.

[10] García, V., Sanchez, J., Mollinenda, R. (2008) "An Empirical Study of the Behavior of Classifiers on Imbalanced and Overlapped Data Sets", Lecture Notes in Computer Science, Volume 4756, Progress in Pattern Recognition, Image Analysis and Applications, Pages 397-406.

[11] Hackshaw, AK and Law, MR and Wald, NJ. 1997. The accumulated evidence on lung cancer and environmental tobacco smoke. British Medical Journal, 315 (7114), 980-988.

[12] Hall M and Eibe F and Holmes G and Pfahringer B and Reutemann P and Witten I H. 2009. The WEKA Data Mining Software: An Update; SIGKDD Explorations, 11 (1).

[13] Kowsalya, R. Sasikala, G, Sangeetha Priya J. 2010. Acute Cystitis and Acute Nephritis Prediction Using Machine Learning Techniques. Global Journal of Computer Science and Technology, 10 (8), 14-16.

[14] Lavesson, Niklas and Davidson, Paul. 2004. A multidimensional measure function for classifier performance. In Proceedings of the 2nd IEEE International Conference on Intelligent Systems, June 2004, pp.508-513.

[15] O'Reilly, Katherine M A and Mclaughlin, Anne Marie and Beckett, William S and Sime, Patricia J. 2007. Asbestos-related lung disease., American Family Physician, 75 (5), 683-8.

[16] Panda M and Patra M R. 2008. A Comparative Study of Data Mining Algorithms for Network Intrusion Detection. In Proceedings of the 2008 First International Conference on Emerging Trends in Engineering and Technology (ICETET '08). IEEE Computer Society, Washington, DC, USA, 504-507.

[17] Panda M and Patra M R. 2007. Network Intrusion Detection using Naive Bayes, International journal of Computer Science and Network Security, Decˆ AZ30-2007, pp.258-263.

[18] Peddabachigari, S., Abraham, A., Grosan, G., Thomas, J.. 2007. Modeling intrusion detection system using hybrid intelligent systems. J. Netw. Comput. Appl. 30, 1 (January 2007), 114-132.

[19] Quinlan R J. 1990. Decision Trees and Decision Making. IEEE Transaction on System Man Cyber, March/April 1990, 20(2).

[20] Quinlan R J. 1993. C4.5: Programs for Machine Learning. Morgan Kaufmann Publishers Inc., San Francisco, CA, USA.

[21] Quinlan R J. 1996. Improved use of continuous attributes in C4.5. Journal of Artificial Intelligence Research, 4:77-90, 1996.

[22] Samet J M and Wiggins C L and Humble C G and Pathak D R. 1998. Cigarette smoking and lung cancer in New Mexico. The American Review of Respiratory Disease

[23] Soman, T. and Bobbie, P.O. 2005, Classification of Arrhythmia Using Machine Learning Techniques. WSEAS Transactions on Computers, Vol.4, issue 6, June 2005, pp. 548-552.

[24] Surveillance, Epidemiology, and End Results (SEER) Program (www.seer.cancer.gov) Research Data (1973-2004), National Cancer Institute, DCCPS, Surveillance Research Program, Cancer Statistics Branch, released April 2010, based on the November 2009 submission.

[25] U.S. Cancer Statistics Working Group. 2010. United States Cancer Statistics: 1999-2006 Incidence and Mortality Web-based Report. Atlanta (GA): Department of Health and Human Services, Centers for Disease Control and Prevention, and National Cancer Institute; 2010. URL: "http://www.cdc.gov/uscs".

[26] Yanwei Xing, Jie Wang, Zhihong Zhao, and and Yonghong Gao. 2007. Combination Data Mining Methods with New Medical Data to Predicting Outcome of Coronary Heart Disease. In Proceedings of the 2007 International Conference on Convergence Information Technology (ICCIT '07). IEEE Computer Society, Washington, DC, USA, 868-872.

[27] Youn S. and McLeod D. 2006. A Comparative Study for Email Classification, Proceedings of International Joint Conferences on Computer, Information, System Sciences, and Engineering (CISSE'06), Bridgeport CT, December 2006.

[28] World Health Organization. 1992-1994. International Statistical Classification of Dis eases and Related Health Problems (ICD-10), 10th Revision. 3 vols. Geneva: WHO.

[29] Witten Ian H and Frank Eibe. 2005. Data Mining: Practical Machine Learning Tools and Techniques. The Morgan Kaufmann Series in Data Management Systems. 2nd Ed. Morgan Kaufmann, 2005. 1988;137(5):1110-3.

**George Dimitoglou** is Associate Professor of computer science at Hood College. He holds a doctorate in computer science from The George Washington University. He is a member of the IEEE, the ACM and the Mathematical Association of America.

**James A. Adams** is Sr. Systems Analyst with Marriott International. He holds a B.S. from George Mason University in Accounting and Management Information Systems a M.S. in computer science from Hood College where he received the 2008 Departmental Scholar Award.

**Carol M. Jim** is a doctoral student in computer science at The George Washington University. She holds a B.A. in mathematics and computer science and a M.S. in computer science from Hood College where she received the 2010 Departmental Scholar Award.